\documentclass[letterpaper]{article} 
\usepackage{aaai24}  
\usepackage{times}  
\usepackage{helvet}  
\usepackage{courier}  
\usepackage[hyphens]{url}  
\usepackage{graphicx} 
\usepackage{natbib}  
\usepackage{caption} 
\usepackage{algorithm}
\usepackage{algorithmic}
\usepackage{amssymb}
\usepackage{amsmath}
\usepackage{booktabs}
\usepackage[table]{xcolor}
\usepackage{subcaption}
\usepackage{float}

\usepackage{newfloat}
\usepackage{listings}
\DeclareCaptionStyle{ruled}{labelfont=normalfont,labelsep=colon,strut=off} 
\floatstyle{ruled}
\newfloat{listing}{tb}{lst}{}
\floatname{listing}{Listing}
\nocopyright

\title{Can the Query-based Object Detector Be Designed with Fewer Stages?}
\author {
    Jialin Li,
    Weifu Fu,
    Yuhuan Lin,
    Qiang Nie,
    Yong Liu
}
\affiliations {
    Tencent Youtu Lab\\
}

\begin{document}

\maketitle

\begin{figure*}[ht]
    \centering
    \begin{minipage}[t]{0.3\textwidth}
        \centering
        \includegraphics[width=0.88\textwidth]{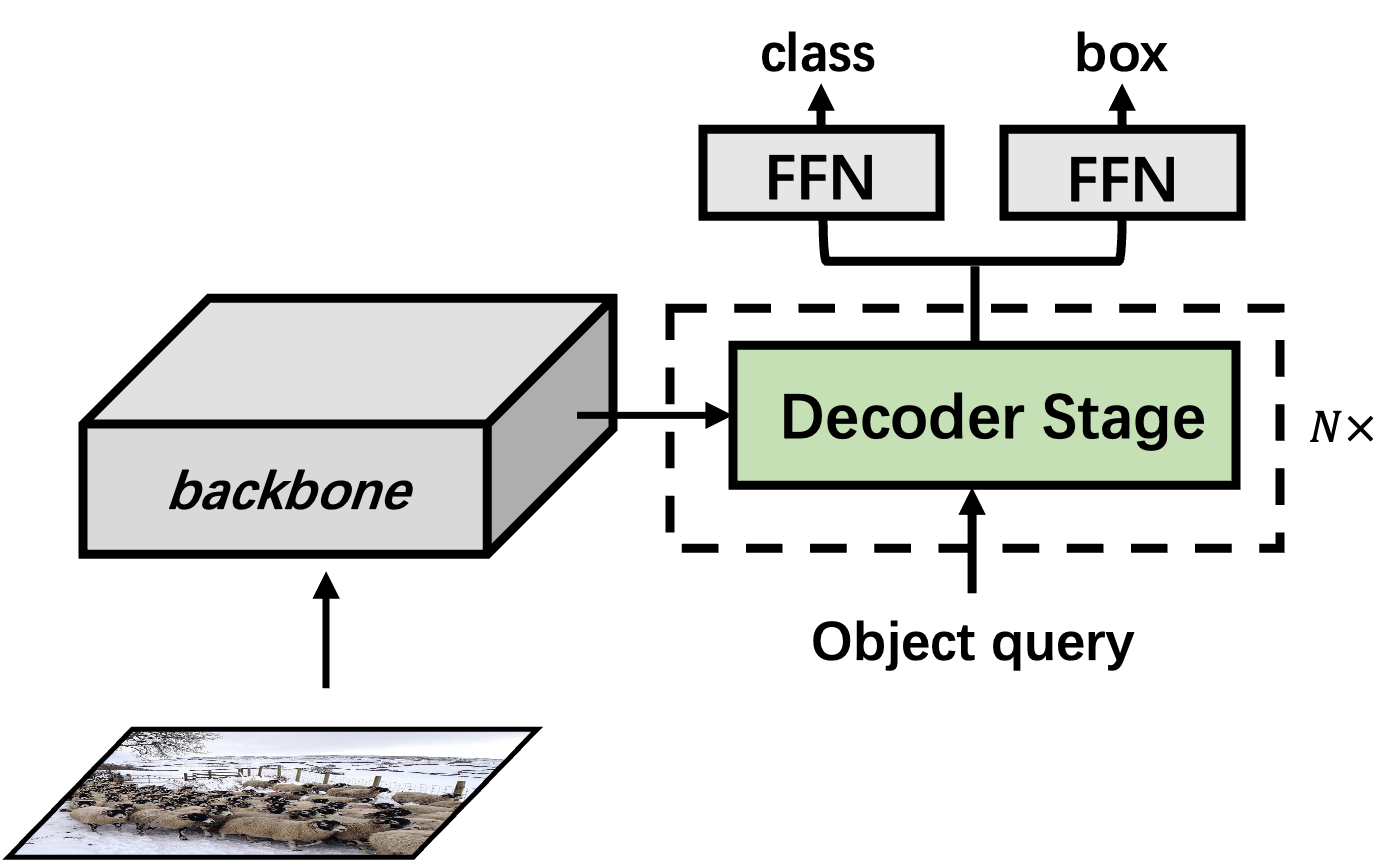}
        \caption*{Plain object query}
        \label{fig1_plain_object_query}
    \end{minipage}%
    \begin{minipage}[t]{0.3\textwidth}
        \centering
        \includegraphics[width=0.88\textwidth]{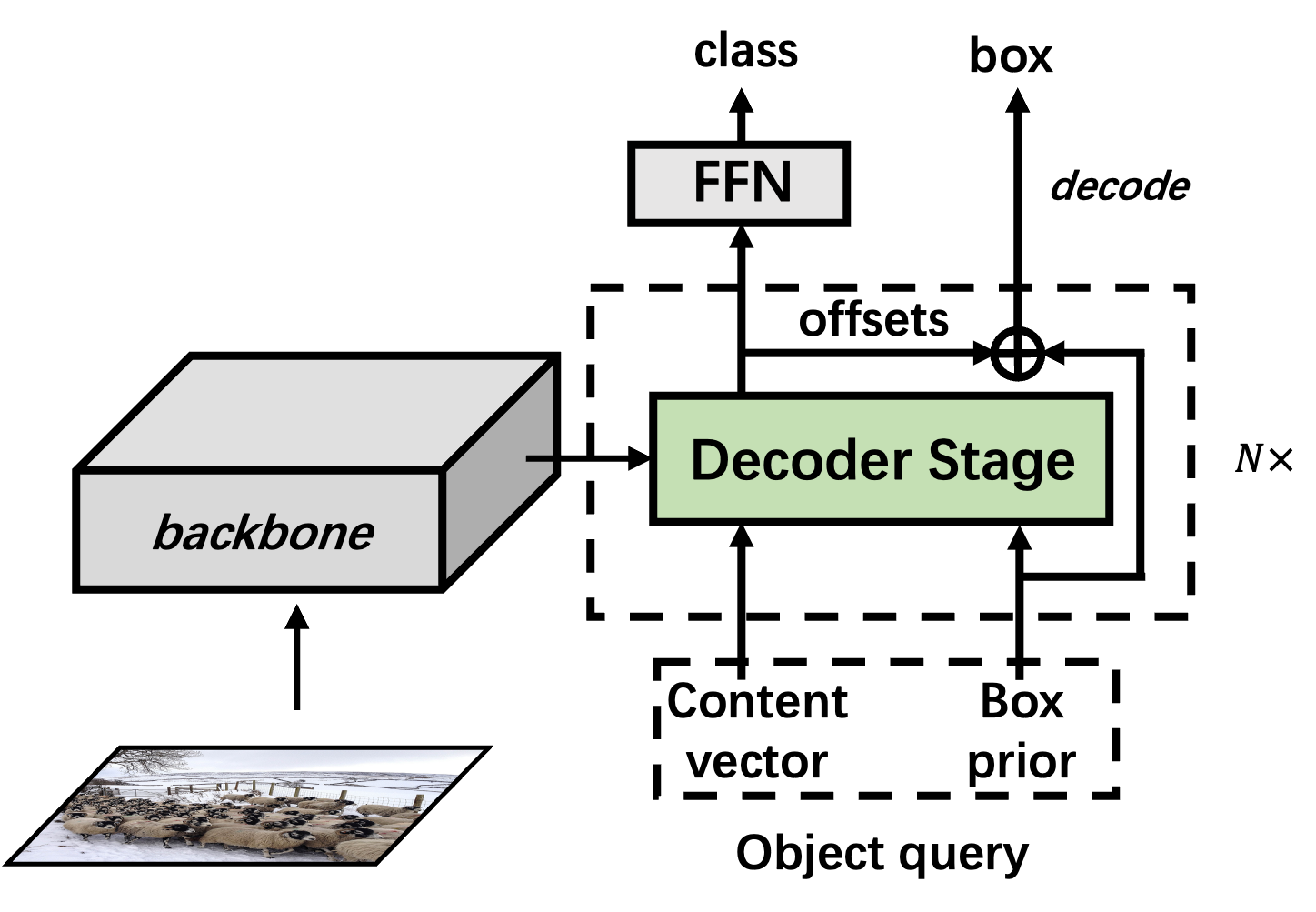}
        \caption*{Decoupled query}
        \label{fig1_decoupled_query}
    \end{minipage}%
    \begin{minipage}[t]{0.3\textwidth}
        \centering
        \includegraphics[width=0.88\textwidth]{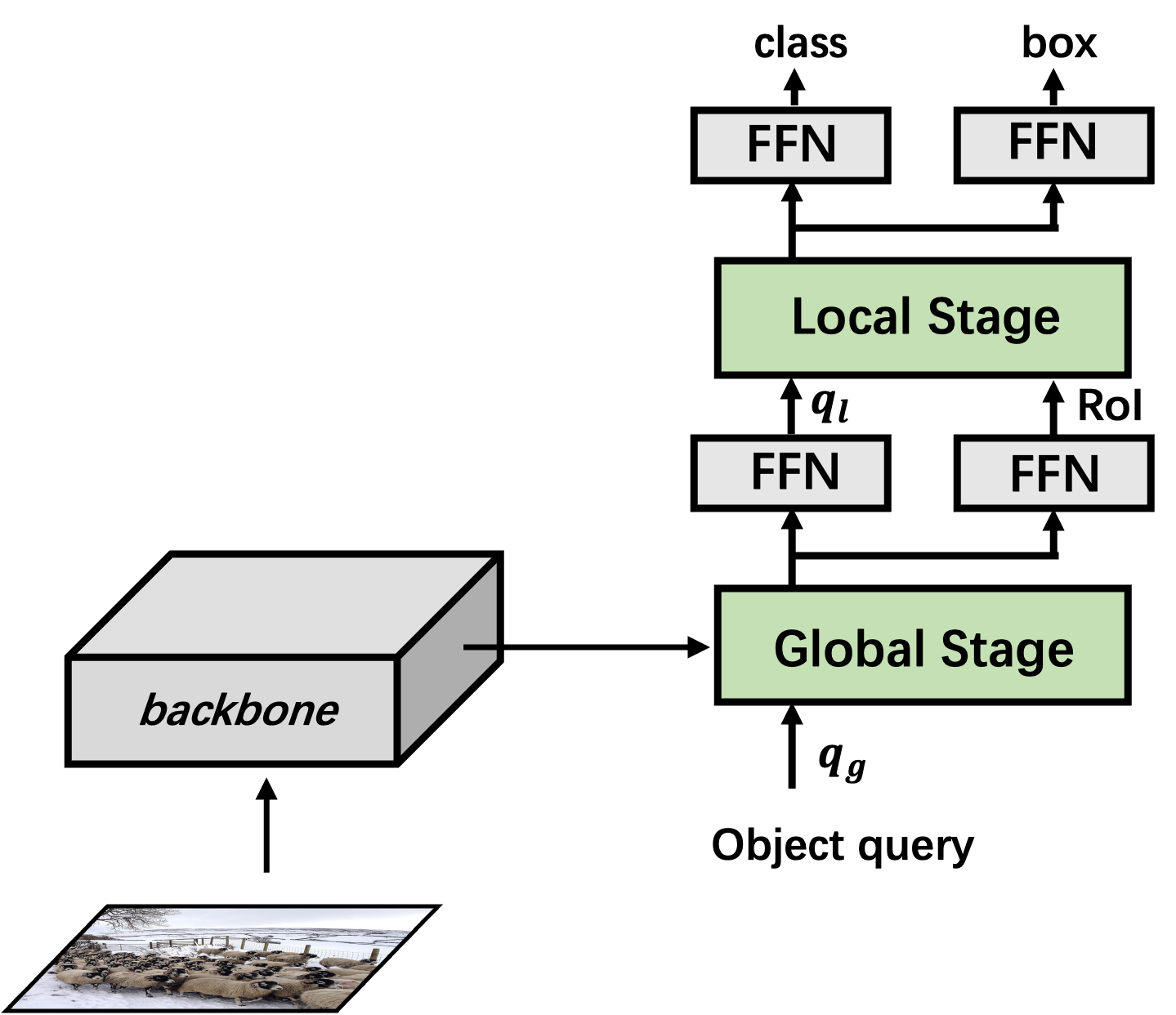}
        \caption*{GOLO}
        \label{fig1_GOLO}
    \end{minipage}
    \caption{Comparisons of different paradigms of query-based methods.}
    \label{fig1}
\end{figure*}

\begin{abstract}
Query-based object detectors have made significant advancements since the publication of DETR. However, most existing methods still rely on multi-stage encoders and decoders, or a combination of both. Despite achieving high accuracy, the multi-stage paradigm (typically consisting of 6 stages) suffers from issues such as heavy computational burden, prompting us to reconsider its necessity. In this paper, we explore multiple techniques to enhance query-based detectors and, based on these findings, propose a novel model called GOLO ($\textbf{G}$lobal  $\textbf{O}$nce and $\textbf{L}$ocal $\textbf{O}$nce), which follows a two-stage decoding paradigm. Compared to other mainstream query-based models with multi-stage decoders, our model employs fewer decoder stages while still achieving considerable performance. Experimental results on the COCO dataset demonstrate the effectiveness of our approach. 
\end{abstract}

\section{Introduction}
Object detection is a crucial task in computer vision, involving the localization and classification of objects in images. Convolutional neural networks (CNNs) \cite{sultana2020review} have demonstrated impressive performance in this field. However, these models still rely on numerous hand-designed components during training and inference, limiting their flexibility and generalizability.

The emergence of query-based detectors, such as DETR \cite{carion2020end}, has introduced a new class of object detectors that eliminate the need for hand-designed components like non-maximum suppression (NMS). These models are characterized by their simplicity and high performance. However, they often heavily rely on multi-stage encoders or decoders, which typically consist of six stages. Among these, multi-stage decoders are considered more crucial than multi-stage encoders, as evidenced by various detectors \cite{carion2020end,zhu2020deformable,meng2021conditional,liu2022dab,sun2021sparse,gao2022adamixer,wang2022anchor,Hong_2022_CVPR}. Although these multi-stage decoders play a crucial role, the inclusion of multiple stages results in increased computational complexity, longer training times, and slower inference. 
However, due to query-based detectors relying on progressive and dynamic refinement throughout training and inference, multi-stage decoders are essential for achieving the desired performance.

Given this challenge, the question arises: how can we reduce the number of stages and improve efficiency in query-based detectors?

In this work, we propose a two-stage solution called GOLO (Global Once and Local Once) to address our concern. The combination of global localization and local refinement shall serve as a direction for eliminating the need for a multi-stage paradigm. As shown on the right side of Figure \ref{fig1}, GOLO first coarsely localizes regions of interest (RoIs) on a global scale and then performs accurate regression and classification for each RoI on a local scale. Each stage incorporates several innovative techniques.

In the global object localization stage, we propose an effective approach to integrate features from multiple scales, enabling each query to subsequently perceive multi-scale features of the entire image at a relatively low cost. Concurrently, we adopt a binary classification scheme to distinguish between foreground and background without considering specific object categories, which accelerates model convergence during early training stages without compromising performance. Moreover, compared to random initialization, we design a meta-based query initialization that enhances the correlation between initialized queries at a minimal cost, thereby improving the training effect. Additionally, we develop a bidirectional adaptive sampling approach, which allows for more effective utilization of scale-aware information and facilitates the global integration of features from the backbone feature maps, which is also employed in the subsequent local stage.

In the local refinement stage, we design a query-guided feature enhancement method that aims to better reinforce the region-based features. Furthermore, we devise an enhanced loss strategy to aid in achieving better model optimization, which is also effective in the preceding global stage. By comprehensively employing these two innovative techniques, we are able to achieve satisfactory detection results with just one local refinement process.

Incorporating all the aforementioned improvements allows us to successfully reduce the decoder stages of our model to only two: a global stage and a local stage. This reduction strikes a favorable balance between computational cost and inference accuracy. Extensive experiments on the COCO dataset \cite{lin2014microsoft} validate the effectiveness of our proposed method.

\section{Related Works}
We first examine the number of stages in various detectors, and then analyze different design paradigms for query-based detectors' queries. 
\subsection{Number of Stages in Detectors} Our research utilizes the matching of model output and
ground truth during training to distinguish between different decoder stages, as the matching result influences the subsequent stage's model state or input.
Accordingly, object detection models are categorized into three groups: one-stage, two-stage, and multi-stage.

\subsubsection{One-stage} These detectors perform object detection in a single shot without region proposal or cascaded decoders \cite{redmon2016you,liu2016ssd,lin2017focal,zhou2019objects,redmon2017yolo9000,tian2019fcos,law2018cornernet}. Examples include YOLO \cite{redmon2016you}, SSD \cite{liu2016ssd}, RetinaNet \cite{lin2017focal}, and CenterNet \cite{zhou2019objects}. YOLO employs a single CNN to predict bounding boxes and class labels of objects. SSD predicts these attributes at multiple scales. RetinaNet addresses class imbalance using a focal loss. CenterNet utilizes keypoint detection to predict object center points and sizes. One-stage detectors offer fast speed and high efficiency; however, they tend to have relatively lower detection accuracy.

\subsubsection{Two-stage} These detectors generate region proposals and then classify and refine these proposals to obtain final detection results. Representative examples include R-CNN \cite{girshick2014rich}, Fast R-CNN \cite{girshick2015fast}, Faster R-CNN \cite{ren2015faster}, and Mask R-CNN \cite{he2017mask}. R-CNN uses selective search for region proposal generation and extracts features using a CNN. Fast R-CNN and Faster R-CNN utilize a region proposal network, which greatly improves speed. Mask R-CNN extends Faster R-CNN by adding a segmentation branch for object mask prediction.

\subsubsection{Multi-stage} These detectors employ multiple encoder or decoder layers to refine detection results. Cascade R-CNN is an early and well-known example. Query-based models largely follow this paradigm. DETR \cite{carion2020end}, the origin of query-based models, uses a transformer-based architecture with 6 encoder and 6 decoder stages to predict object sets and their bounding boxes. Other models like Sparse R-CNN \cite{sun2021sparse}, Deformable DETR \cite{zhu2020deformable}, AdaMixer \cite{gao2022adamixer}, DAB-DETR \cite{liu2022dab} and DINO \cite{zhang2022dino} introduce various techniques to improve detection accuracy, reduce training time, and enhance model adaptability. 

\subsection{Query Design Paradigms}
At present, there are mainly three design paradigms for interpreting and initializing queries in query-based object detectors, as illustrated in Figure \ref{fig1}. 

\subsubsection{Plain Object Query} The first paradigm decodes object classification and bounding boxes from the final output query after a series of operations applied to randomly initialized queries, similar to the original DETR architecture \cite{carion2020end,zhu2020deformable,gao2021fast}. 

\subsubsection{Decoupled Query} The second paradigm separates the content vector and spatial vector from the query, optimizes them simultaneously, and then uses them to decode object classification and bounding boxes separately \cite{sun2021sparse,gao2022adamixer,wang2022anchor,zhang2022dino,meng2021conditional}. While this design reduces optimization difficulty and improves model performance, its spatial vector initialization is usually independent of the image and requires multiple iterations to approach the ground truth. There exist methods that attempt to select initial queries (or spatial vector) from dense features \cite{zhang2022featurized,yao2021efficient}, but they either introduce hand-designed components or require larger computational resources.

\subsubsection{Our Approach}We provide a different design for queries compared to the first two paradigms. We first enable the query to distinguish foreground and background and estimate approximate object bounding boxes using a global stage. Then, we allow the query to complete region-based detailed classification and regression using a local stage. This approach aids in achieving better training convergence and aligns with the traditional two-stage object detection paradigm \cite{girshick2014rich,girshick2015fast,ren2015faster,he2017mask}.

\section{The Proposed Method}
Our proposed model, GOLO, adheres to the query-based paradigm while seeks to avoid using excessive stages. It tries to achieve comparable performance with only two decoder stages, one global and one local.

\subsection{The Overall Pipeline}
\begin{figure}[H]
  \centering
  \includegraphics[width=\linewidth]{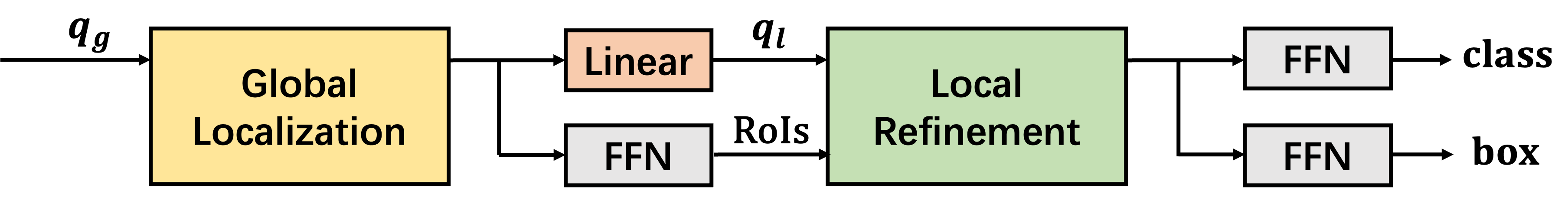}
  \caption{Overview of our architecture. During the global localization stage, the query is denoted as $q_g$, while during the local refinement stage, the query is denoted as $q_l$. FFN stands for Feedforward Network.}
  \label{over-view-pipeline}
\end{figure}

As depicted in Figure \ref{over-view-pipeline},
our model operates entirely in a query-based manner. It consists of two main stages: global object localization and local feature-guided refinement.
In the first stage, global object localization is performed to generate a set of candidate Region of Interest (RoIs), which helps avoid the need for a cascaded decoder caused by the randomness of query initialization.
In the second stage, RoIs are utilized to obtain locally distinctive features for refining the final classification and bounding box regression. 

\subsection{Global Localization}
\label{Global Localization}

\subsubsection{The Detailed Main Structure}
\begin{figure}[htbp]
\centering
\begin{minipage}[t]{\linewidth}
\centering
\includegraphics[width=7cm]{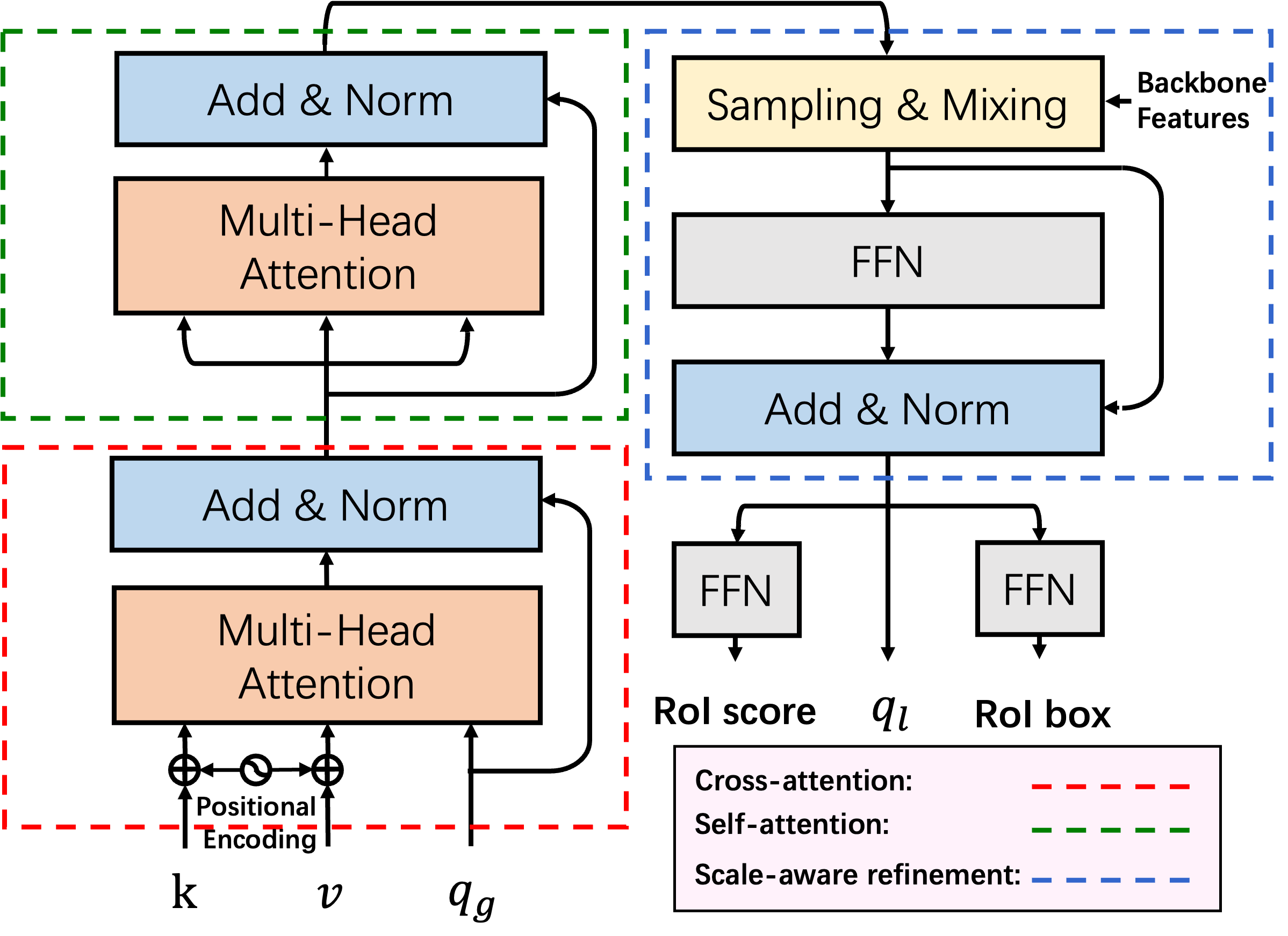}
\caption{Global localization stage}
\label{global localization stage fig}
\end{minipage}
\end{figure}

This stage consists of the following main components: cross-attention, self-attention, point-wise feature sampling and mixing. All these components are illustrated in Figure \ref{global localization stage fig}. Cross-attention and self-attention employ the multi-head attention mechanism proposed in \cite{vaswani2017attention}. To facilitate point-wise feature sampling and mixing, we begin by employing our proposed bidirectional adaptive sampling method to obtain weighted features. A comprehensive introduction to this method will be provided later on.
Then, we perform adaptive channel mixing and adaptive spatial mixing, as proposed in \cite{gao2022adamixer}, with parameter settings aligned with the original papers. Detailed descriptions of multi-head attention, adaptive channel mixing, and adaptive spatial mixing can be found in \cite{vaswani2017attention,gao2022adamixer}.

It is necessary to provide a detailed explanation of how our cross-attention is implemented, as it enables each query to perceive a global information of the image and locate its own corresponding RoI. We utilize object queries $\mathbf{Q}$ and features $\mathbf{K}$ and $\mathbf{V}$ from the image, where $\mathbf{K}$ and $\mathbf{V}$ come from the multi-scale feature fusion aforementioned. The cross-attention operation can be formulated as follows:
\begin{equation}
\text{Cross-Attention}(\mathbf{Q}, \mathbf{K}, \mathbf{V}) = \text{softmax}\left(\frac{\mathbf{Q}\mathbf{K}^\top}{\sqrt{d_k}}\right)\mathbf{V}
\end{equation}
where $\mathbf{Q}$ is a matrix consisting of n different object queries $\mathbf{q_i}$, $i \in {1, 2, \dots, n}$, $n$ is the total number of object queries. $\mathbf{K}$ and $\mathbf{V}$ are obtained from a linear transformation of the multi-scale feature fusion output $X_{mff} \in \mathbb{R}^{h_s \times w_s \times c \times k_{mff}}$. The linear transformation is applied to the last dimension of $X_{mff}$, resulting in a variable $Y \in \mathbb{R}^{h_s \times w_s \times c \times 2}$.  The last dimension of $Y$ is then separated, and the first and second dimensions of the separated variable are flattened to obtain $\mathbf{K}$ and $\mathbf{V}$ respectively, with the same shapes as $(h_s*w_s, c)$.

Following cross-attention, the subsequent step is self-attention, which facilitates the exchange of information among queries. This mechanism empowers each query to identify the most suitable match and make informed classification decisions.
The combination of cross-attention and self-attention mechanisms provides the query with the necessary information for both localization and classification, enabling it to focus more effectively on the relevant regions of the input. To further enhance the query's ability to extract and represent information, we perform a round of point-wise feature sampling and mixing as the last step in this stage. This allows each query to extract features from the entire image and fuse them with its own features, improving the accuracy of localization and foreground/background discrimination. We demonstrate the effectiveness of this module through experiments, as shown in Table \ref{table_scale_aware_refinement}.

\subsubsection{Meta-based Query Initialization}
We assume that object queries should implicitly encompass combinations of fundamental conditions regarding objects, such as their essential visual features (e.g., color, shape, size, orientation, etc.). These unique combinations vary across different queries and are responsible for distinguishing different objects. Based on the aforementioned considerations, we propose the introduction of a set of "meta" vectors that represent the fundamental features mentioned earlier. By doing so, object queries can be expressed as linear combinations of these "meta" vectors. Specifically, we introduce a set of $m$ learnable "meta" vectors of dimension $c$, which represent the basic visual features of objects. We set $m=256$, $c=256$ by default. These "meta" vectors are linearly combined into object queries. More formally, let 
$\mathbf{q} \in \mathbb{R}^{n \times c}$ denote the object queries, we can express each query $\mathbf{q}_i$ as a linear combination of $m$ "meta" vectors $\mathbf{m}$:
\begin{equation}
\mathbf{q}_i = \sum_{j=1}^{m} \alpha_{ij} \mathbf{m}_j
\end{equation}

where $\alpha_{ij}$ is the weight assigned to the $j$-th "meta" vector for the $i$-th query. The "meta" vectors $\mathbf{m}$ and weights $\alpha$ are both learnable during training. Experiments show that this initialization results in improved representation of object features during training and leads to a better model performance, which is demonstrated in Table \ref{table_CA_Meta}.

\subsubsection{Multi-scale Feature Fusion}
\begin{figure}[t]
\centering
\includegraphics[width=\linewidth]{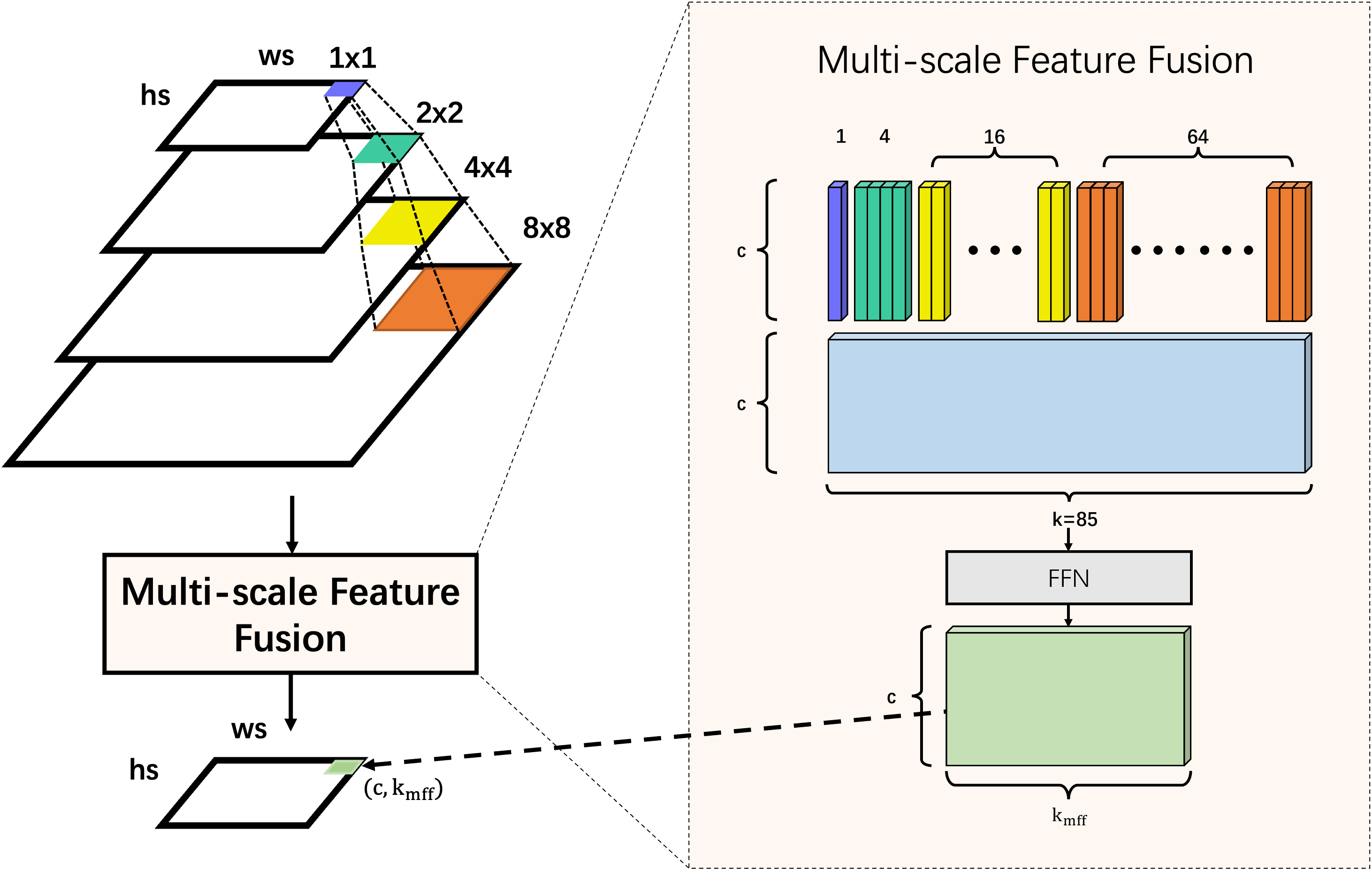}
\caption{An overview of multi-scale feature fusion. $(h_s, w_s)$ is defined as the size of the smallest feature map output from the backbone. Then, for each element on the smallest feature map, we extract corresponding features from the same position and aligned scale on the other three larger feature maps, and all these four layer features are concatenated into a $(c, 85)$ shape. Next, the vector is passed through two linear transform layers to first expand and then reduce the last dimension to obtain a $(c, k_{mff})$ vector. By iterating through all the elements on the smallest feature map, we obtain an MFF featurre of size $(h_s, w_s, c, k_{mff})$. This process enables the MFF to capture multi-scale contextual information and enhance the discriminative power of the features for object detection.}
\label{multi-scale feature fusion}
\end{figure}

We use this module to aggregate different scale features from the backbone for a rough global object localization of the image.
ResNet architecture \cite{he2016deep} is utilized as the backbone to generate multi-scale feature maps from the input image. We incorporate $P_2$ to $P_5$ levels in our model, Each of which has a resolution half of the previous level and consists of 256 channels.
All feature maps are channel-wise transformed to the same size of the smallest one to generate a global feature map, with a size of $(h_s, w_s, c, k_{mff})$, where $h_s$ and $w_s$ are the height and width of the smallest feature map, $c$ is the number of channels, typically 256, and $k_{mff}$ represents the number of kernels outputted at the same position on the feature map, which is set to 64 by default. The detailed process is depicted as Figure \ref{multi-scale feature fusion}.

Please note that our approach differs from the methods proposed in Featurized Query R-CNN\cite{zhang2022featurized} or DINO \cite{zhang2022dino} . In those works, they attempt to define rules based on selected features from feature maps. These selected features are then used to initialize the query using certain initialization techniques. This process involves several manually designed components, such as threshold selection or top-k point selection. In contrast, our approach aims to globally perceive the overall image features directly through the aforementioned method. Subsequently, through training in the global stage, the query is automatically able to distinguish foreground regions. The entire process does not involve any manual components. Considering that one major advantage of query-based detectors is their simplicity in design and the robustness they bring in practical applications, we believe that our design aligns better with the development trend.

\subsubsection{Bidirectional Adaptive Sampling}
\label{Bidirectional Adaptive Sampling}
After self-attention in the global or local stage, we perform adaptive feature sampling for each query on the backbone feature maps based on its own features, and then fuse the sampled features to enhance the query. The fusion process uses adaptive channel mixing and adaptive spatial mixing proposed in \cite{gao2022adamixer}. But considering sampling method used in \cite{gao2022adamixer}, it may not be as suitable for objects with significant differences in aspect ratio. In such cases, it is important to consider a more reasonable approach for sampling multi-scaled features.

Assuming a bounding box for an object has width and height $(\Tilde{h},\Tilde{w})$. In a bad case, the scales of $\Tilde{h}$ and $\Tilde{w}$ can differ significantly, such as when $\tilde{w} \gg \tilde{h}$. In such cases, different scales of features should be selected in the $\Tilde{w}$ and $\Tilde{h}$ directions. Therefore, we propose bidirectional adaptive sampling to sample features based on the bidirectional scale information, which is generated by the query.

Specifically, we denote the z-axis coordinate of a feature map as $z^{feat}_j = log_2(s^{feat}_j)$, where $s^{feat}_j$ is the downsampling stride of the $j$-th feature map.
Let $(x_i, y_i, z_i^w, z_i^h)$ be a linear transformation $T_i$ of a query $q$, 
\begin{equation}
(x_i, y_i, z_i^w, z_i^h) = T_i(q), i \in{1, 2, \ldots, n}
\end{equation}
where $x_i$ and $y_i$ are the planar coordinates of a point to be sampled by the query, $z_i^w$ and $z_i^h$ are its coordinates in the z-axis direction respectively with regard to $w$ and $h$ direction, and $n$ is the number of points being sampled.
Then our weighted sampling can be expressed as follows: 
\begin{equation}
f^{i_{sampled}} = \sum_{j=1}^{K} w_j * f_{j}(x_i,y_i)
\end{equation}
where $K$ is the total number of feature maps. $w_i$ is the weight assigned to the $i$-th feature map calculated by a Gaussian function with a standard deviation $2$:

\begin{equation}
\begin{aligned}
w_j = \frac{\exp\left(-(z_j^h - \tilde{z}^{h}_j)^2/2\right)}{\sum_{k=1}^{K}\exp\left(-(z_k^h - \tilde{z}^{h}_j)^2/2\right)} \\
+ \frac{\exp\left(-(z_j^w - \tilde{z}^{w}_j)^2/2\right)}{\sum_{k=1}^{K}\exp\left(-(z_k^w - \tilde{z}^{w}_j)^2/2\right)}
\end{aligned}
\end{equation}

where $f_{j}(x_i, y_i)$ is the feature value at the point $(x_i, y_i)$ in the $j$-th feature map, $f^{i_{sampled}}$ is the weighted sampled features.
By using this bidirectional adaptive sampling method, we assert that the obtained sampling features $\{f^{1_{sampled}}, f^{2_{sampled}}, \ldots, f^{3_{sampled}}\}$ are more suitable for query $q$, which is further supported by the results obtained in experiment  shown in Table \ref{table_bidirectional_adaptive_sampling}.

\subsection{Local Refinement}
\label{Local Refinement}
\subsubsection{The Detailed Structure}
\begin{figure}[htbp]
\centering
\centering
\includegraphics[width=7cm]{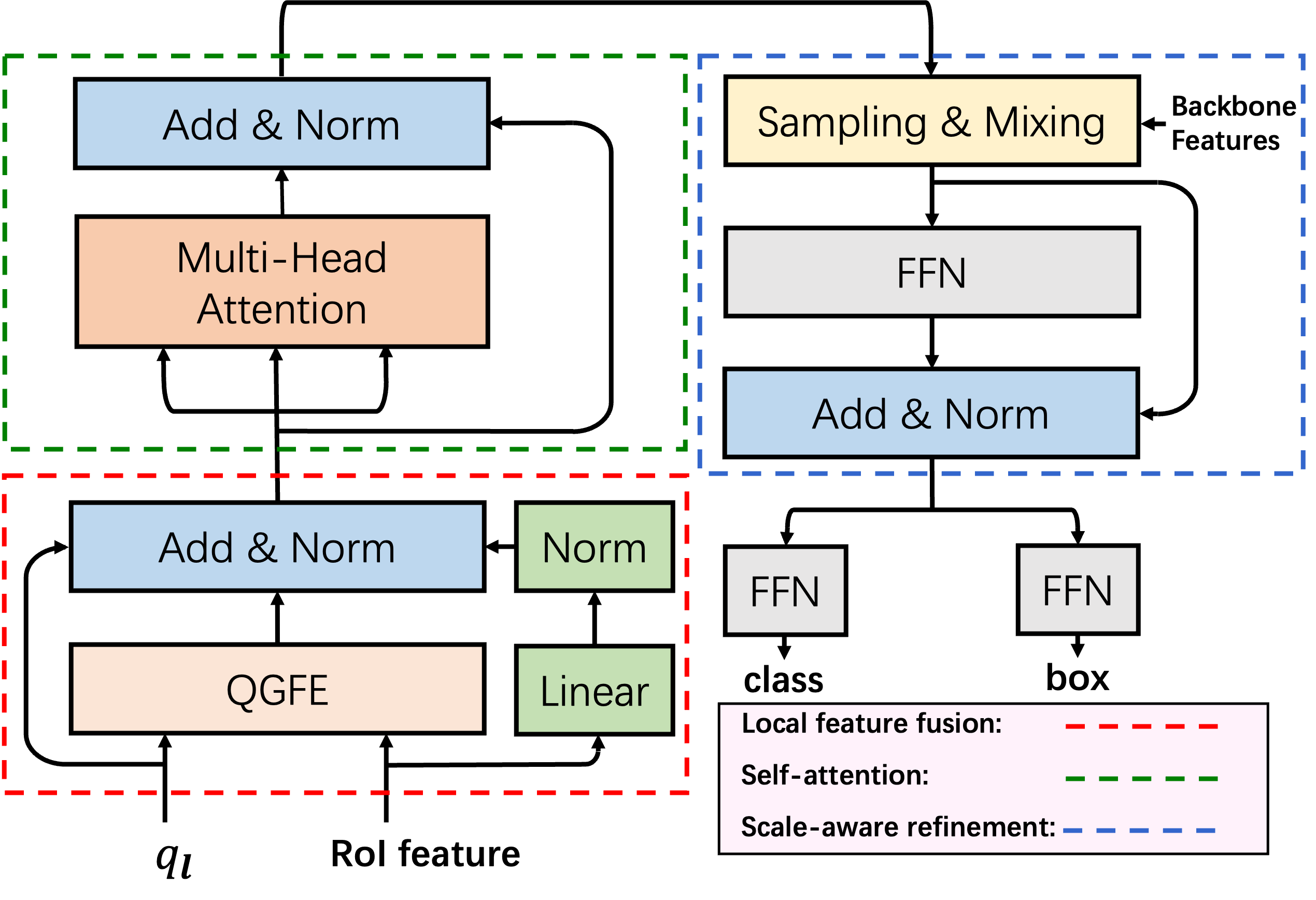}
\caption{Local refinement stage}
\label{local refinement stage fig}
\end{figure}

The local refinement stage also comprises three stages: query-guided feature enhancing (QGFE), self-attention, and point-wise feature sampling and mixing, as shown in Figure \ref{local refinement stage fig}. In the first stage, RoI boxes $B$ obtained from the previous stage are used to extract image features within each RoI from backbone features. We use RoIAlign \cite{he2017mask} to accomplish this step. The extracted features are then interacted and fused with the object queries $\mathbf{Q_g}$ from the previous stage to incorporate the RoI features into the queries.
After feature fusing, similar to the global localization stage, the queries in the local refinement stage undergo a self-attention mechanism to exchange information, which is primarily aimed at improving the classification performance. Then a round of point-wise feature sampling and mixing is performed to reinforce the information contained in the queries. Finally, the classification and bounding box regression for each query are obtained using a classification head and a box head respectively. Head settings are identical to Faster R-CNN \cite{ren2015faster}.

\subsubsection{Query-Guided Feature Enhancing}
Formally, for a given query $q_i$, its corresponding feature from RoIAlign is denoted as $f_i$, which has a dimension of $(S*S, C)$, let $d_i$ be the query-guided feature enhancing of $f_i$ with regard to $q_i$,
\begin{equation}
d_i = \text{QGFE}(q_i, f_i)
\end{equation}
where the computation process of QGFE is elaborated in detail in Algorithm \ref{alg:Query-guided RoI feature enhancing}, considering a pair of query and its RoI feature. It adaptively enhances the ROI features based on the information from the query.
Let $r_i$ be the result of applying two linear transformations to $f_i$, within which is a layer normalization\cite{ba2016layer} and a ReLU activation:
\begin{equation}
r_i = W_2(LN(ReLU(W_1(f_i))))
\end{equation}
where $W_1$ and $W_2$ are two linear transformations, $LN$ denotes layer normalization, and $\text{ReLU}$ denotes rectified linear unit\cite{agarap2018deep}.
The region-based fused query $\hat{q_i}$ is obtained by adding $q_i$, $d_i$, and $r_i$ element-wise, i.e., 

\begin{equation}
\hat{q_i} = q_i + d_i + r_i
\end{equation}

\begin{algorithm}[H]
\caption{Query-guided feature enhancing}
\label{alg:Query-guided RoI feature enhancing}
\textbf{Input}: {$\mathbf{q_g}: (1, C)$; $\mathbf{roi\_feats}: (S * S, C)$}\\
\begin{algorithmic}[1] 
\STATE $\mathbf{q_g} = \mathbf{q_g}.\text{flatten}()$
\STATE $K_1, K_2 \gets \text{linear1}(\mathbf{q_g}),\text{linear2}(\mathbf{q_g})$ \\\textcolor{gray}{Note: $K_1$ and $K_2$ have shape $S \times S \times C$.}
\STATE $K_1 = K_1.\text{view}(C, S \times S)$\\$K_2 = K_2.\text{view}(C, S \times S)$
\STATE $\mathbf{roi\_feats} \gets \text{relu}(\text{norm}(\text{bmm}(\mathbf{roi\_feats}, K_1)))$\\\textcolor{gray}{Note: $roi\_feats$ now have shape $(S \times S, S \times S)$.}
\STATE $\mathbf{roi\_feats} \gets \text{relu}(\text{norm}(\text{bmm}(K_2, \mathbf{roi\_feats})))$\\\textcolor{gray}{Note: $roi\_feats$ now have shape $(C, S \times S)$.}
\STATE $\mathbf{feats\_enhanced} \gets \text{linear3}(\mathbf{roi\_feats}.\text{flatten}())$\\\textcolor{gray}{Note: $feats\_enhanced$ is the linear transformation of $roi\_feats$ from $(C \times S \times S)$ to $(C)$.}\\
\end{algorithmic}
\textbf{return}: $\mathbf{feats\_enhanced}.\text{view}(1, C)$
\end{algorithm}

\subsection{Loss Enhancement Strategy}
Our model employs a matching loss \cite{carion2020end} in both global and local stages for training supervision, defined as:

$$L_{match} = \lambda_{cls} L_{cls} + \lambda_{L1} L_{L1} + \lambda_{giou} L_{giou}$$

with $L_{cls}$, $L_{L1}$ and $L_{giou}$ representing classification loss, $L1$ regression loss and generalized intersection over union (GIoU) losses, respectively. 
Hyperparameters $\lambda_{cls}$, $\lambda_{L1}$, and $\lambda_{giou}$ control their relative importance.

We also use auxiliary losses, $L_{aux\_bbox}$ and $L_{aux\_cls}$, in both stages for object localization and classification. We considered three steps in each stage: cross-attention (or feature fusion) with image features, self-attention without image interaction, and scale-aware refinement with sampling. It is important to note that the second step, self-attention, does not rely on image feature interaction, which means that the queries should already possess the information to locate objects after the first step. Additionally, the main purpose of the self-attention is for the queries to communicate with each other, determining the best match and making classification decisions for each query, not object localization.

Inspired by this, we introduced auxiliary bounding box loss and classification losses in the first and second steps, respectively, for both stages. Ground truth values for these losses are determined by the final matching relationship in each stage, ensuring consistent localization and classification. This method not only accelerates the convergence speed during training but also improves the final performance by 0.2 AP.

The total loss $L$ is a linear combination of the matching loss and the auxiliary losses:

$$L = L_{match} + \lambda_{aux\_bbox} L_{aux\_bbox} + \lambda_{aux\_cls} L_{aux\_cls}$$

where $\lambda_{aux\_bbox}$ and $\lambda_{aux\_cls}$ are hyperparameters that control the relative importance of each loss term.

\section{Experiments}
In this section, we will describe the experimental setup and provide implementation details.

\begin{table*}[htb]
\centering
   \begin{tabular}{l|c|c|c|cccccc}
    \toprule
    Method & backbone & Epochs & $N_{decoder}$ & $AP$ & $AP_{50}$ & $AP_{75}$ & $AP_{s}$ & $AP_{m}$ & $AP_{l}$ \\
    \midrule
    RetinaNet & R-50 & 36 & 1 & 39.5 & 58.8 & 42.2 & 23.8 & 43.2 & 50.3 \\
    Faster R-CNN & R-50 & 36 & 2 & 40.3 & 61.0 & 44.4 & 24.0 & 44.1 & 51.4 \\
    Sparse R-CNN & R-50 & 36 & 2 & 33.6 & 50.3 & 36.0 & 16.3 & 35.3 & 47.1\\
    AdaMixer & R-50 & 36 & 2 & 40.5 & 59.5 & 43.7 & 24.0 & 44.1 & 52.6 \\
    GOLO & R-50 & 36 & 2 & \textbf{42.8} & \textbf{61.2} & \textbf{46.2} & \textbf{24.9} & \textbf{45.9} & \textbf{57.0} \\
    \midrule
    RetinaNet & R-101 & 36 & 1 & 41.0 & 60.3 & 44.0 & 24 & 45.3 & 53.8 \\
    Faster R-CNN & R-101 & 36 & 2 & 42.0 & 62.5 & 45.9 & 25.2 & 45.6 & 54.6 \\
    GOLO & R-101 & 36 & 2 & \textbf{44.4} & \textbf{62.8} & \textbf{48.2} & \textbf{27.3} & \textbf{47.8} & \textbf{58.1} \\
    \bottomrule
  \end{tabular}
  \caption{Comparison of detectors with up to 2 decoder stages on the COCO \textit{val2017} dataset. $N_{decoder}$ denotes the number of decoder stages. Owing to the strong baseline performance of Sparse R-CNN and AdaMixer, the limited reported performance of query-based detectors using only two decoder stages, and the constraints on experiment time, we did not compare numerous other detectors in this table.}
  \label{table_main1}
\end{table*}

\subsection{Implementation Details}
\subsubsection{Dataset}
We performed extensive experiments on the widely-used MS COCO\cite{lin2014microsoft} benchmark. All models were trained on the COCO \textit{train2017} split and then evaluated on the \textit{val2017} split.

\subsubsection{Training Details}
We set ResNet architecture as the default backbone, with the AdamW\cite{loshchilov2017decoupled} optimizer and weight decay set to 0.0001. When training with 8 GPUs, we set the batch size to 16 and the learning rate to $2.5 \times 10^{-5}$. All models are trained for a total of 36 epochs, with the learning rate divided by 10 at epochs 27 and 33. The backbone is initialized with an ImageNet-1k\cite{deng2009imagenet} pre-trained model, while the remaining parameters are initialized using Xavier \cite{glorot2010understanding}. Following \cite{carion2020end,sun2021sparse,gao2022adamixer}, we employ random horizontal flips, random cropping, and multi-scale techniques as data augmentation. In particular, the multi-scale techniques makes the shortest side at range of 480 to 800, and the longest side at most 1333. During training, we utilize a matching loss function that consists of focal loss\cite{lin2017focal} with a coefficient of $\lambda_{cls}$ = 2, L1 bounding box loss with $\lambda_{L1}$ = 5 and GIoU\cite{rezatofighi2019generalized} loss with $\lambda_{giou}$ = 2. After conducting preliminary experiments, $\lambda_{aux\_bbox}$ and $\lambda_{aux\_cls}$ are both set to 0.25.

\subsubsection{Model Recipes}
Our model employs only two stages, one global and one local, allowing for a direct comparison with single-stage or two-stage detectors such as RetinaNet\cite{lin2017focal} and Faster R-CNN\cite{ren2015faster}. For comparison with other query-based detectors, we chose Sparse R-CNN \cite{sun2021sparse} and Adamixer \cite{gao2022adamixer} as baselines due to their strong performance, recent publication and decoder-only structure, ensuring a fair comparison with our method. Though our model may achieve better performance with more stages, we restrict the comparison to two stages for all query-based detectors to concentrate on obtaining improved detection results with fewer stages.
We note that the original paper for the recent state-of-the-art detector DINO \cite{zhang2022dino} discussed performance using only two decoder stages. However, DINO's experiments include two additional encoder stages, 100 regular queries and 100 DN queries, and 12 training epochs (compared to our 36 epochs), making a direct and fair performance comparison unfeasible. As such, we do not advise a direct comparison. Nonetheless, we report that under these settings, DINO achieves 41.2 AP, while our model attains 42.8 AP.

\subsection{Main Results}
A performance comparison of our proposed GOLO against other traditional or query-based models is shown in Table \ref{table_main1}. 
To ensure a fair comparison, we set a constraint on the number of decoder stages, permitting a maximum of two stages. This restriction is consistently applied across various models. In Faster R-CNN \cite{ren2015faster}, the two stages are represented by the Region Proposal Network (RPN) and Fast RCNN \cite{girshick2015fast}. Likewise, in Sparse RCNN \cite{sun2021sparse} and Adamixer \cite{gao2022adamixer}, we explicitly set the number of iterative decoder stages to two.
In our model, GOLO, the decoder consists of one global localization stage and one local refinement stage.

Our experimental results demonstrate that our proposed model achieves a high AP of 42.8, the best performance when the number of decoder stages is limited to 2. This proves the effectiveness of our strategy of first performing global localization followed by local refinement for query-based object detectors.
Specifically, GOLO surpasses two-stage Sparse R-CNN and Adamixer by 8.2 AP and 2.3 AP respectively. Notably, with a ResNet-50 backbone, GOLO outperformed the classic Faster R-CNN using ResNet-101. These findings highlight the effectiveness of our overall design in scenarios with limited decoder stages, which is exactly the focus of our work.



\subsection{Ablation Studies}
In this section, we analyze the key innovations in GOLO. We use ResNet-50 as the backbone network and train the models for 36 epochs in all experiments.

\begin{table*}[ht]
\centering
\begin{tabular}{c|cccccc|cccc}
\toprule
Queries & $AP$ & $AP_{50}$ & $AP_{75}$ & $AP_{s}$ & $AP_{m}$ & $AP_{l}$ & FPS & Params(M) &Flops(G) &Training time   \\
\midrule
100	& 39.7 & 57.7 & 42.5 & 22.3 & 42.5 & 53.4 & 20 & 83.2& 157.2& 24h \\
\rowcolor{gray!20} 300 & 42.8 & 61.2 & 46.2 & 24.9 & 45.9 & 57.0 & 19 & 83.3& 168.4& 27h \\
500	& 43.2 & 61.7 & 46.7 & 25.1 & 46.2 & 57.8 & 17 &83.3 &179.5 & 29h \\
\bottomrule
\end{tabular}
\caption{Effect of number of queries. 300 queries strikes the best balance between precision and speed. We use ResNet-50 as the backbone. FPS is evaluated on single NVIDIA Tesla V100 GPU. Training time is calculated on 8 NVIDIA Tesla V100 GPUs. Default choice for our model is colored gray.}
\label{table_queries}
\end{table*}

\subsubsection{Number of Queries}
We examine the effect of varying query quantities on GOLO to find a balance between computational cost and performance. As shown in Table \ref{table_queries}, increasing the number of queries from 100 to 300 led to a 3.1 AP increase, but further increasing it to 500 did not result in a substantial improvement. Therefore, we adopt 300 queries as the default configuration.

\subsubsection{Effect of Meta-based Query Initialization}
Unlike previous query-based detectors that initialize queries (or content vectors) completely randomly, GOLO uses a set of "meta" vectors for query initialization. We examine the effect of meta-based query initialization in Table \ref{table_CA_Meta} and find that incorporating "meta" vectors leads to a 1.3 AP improvement.

\subsubsection{Effect of Multi-Scale Feature Fusion}
During the global localization stage, we implemented a multi-scale fusion mechanism for the backbone features. Before conducting self-attention, we allowed the queries to interact with the fused features through cross-attention, enabling each query to perceive global image features and roughly locate their corresponding RoIs. As demonstrated in Table \ref{table_CA_Meta}, this design leads to a 1.4 AP improvement.

\begin{table}[H]
      \centering
      \begin{tabular}{ccccc}
        \toprule
        MFF & Meta Init & $AP$ & $AP_{50}$ & $AP_{75}$  \\
        \midrule
        \checkmark & & 41.4 & 59.4 & 44.3  \\
        & \checkmark & 41.5 & 59.6 & 44.5  \\
        \rowcolor{gray!20} \checkmark& \checkmark& 42.8 & 61.2 & 46.2  \\
        \bottomrule
        \end{tabular}
        \caption{Effect of Cross Attention and Meta-based Query Initialization.}
        \label{table_CA_Meta}
\end{table}

\subsubsection{Effect of Bidirectional Adaptive Feature Sampling}
We incorporate bidirectional adaptive feature sampling in both the global localization and local refinement stages, which we believe is a more scale-aware sampling method. Table \ref{table_bidirectional_adaptive_sampling} shows a 0.3 AP increase, demonstrating its effectiveness.

\begin{table}[H]
      \centering
      \begin{tabular}{cccc}
        \toprule
        BAFS & $AP$ & $AP_{50}$ & $AP_{75}$  \\
        \midrule
         & 42.5 & 60.8 & 45.6  \\
         \rowcolor{gray!20} \checkmark &  42.8 & 61.2 & 46.2  \\
        \bottomrule
        \end{tabular}
        \caption{Effect of bidirectional adaptive feature sampling. Line with no \checkmark means we use the original sampling methods in AdaMixer.}
        \label{table_bidirectional_adaptive_sampling}
    \end{table}

\subsubsection{Comprehensive Effect of Feature Sampling and Mixing}
Here, we comprehensively examine the combined effect of adaptive feature sampling and subsequent feature mixing. Both global and local stages employ this process before completion, collectively leading to a 6.4 AP improvement as shown in Table \ref{table_scale_aware_refinement}.

\begin{table}[H]
      \centering
      \begin{tabular}{ccccc}
        \toprule
        Global & Local & $AP$ & $AP_{50}$ & $AP_{75}$  \\
        \midrule
        &   & 36.4 & 54 & 39.1  \\
        \checkmark & & 39.0 & 57.5 & 41.9  \\
        & \checkmark & 40.0 & 57.6 & 43.1  \\
        \rowcolor{gray!20} \checkmark& \checkmark& 42.8 & 61.2 & 46.2  \\
        \bottomrule
        \end{tabular}
        \caption{Effect of adaptive feature mixing in global and local stage. \checkmark indicates the usage of this module.}
        \label{table_scale_aware_refinement}
    \end{table}

\section{Conclusion}
In this work, we have presented a two-stage query-based detector, named GOLO, which attains satisfactory performance through a global localization stage and a local refinement stage. Notably, GOLO not only demonstrates comprehensive advantages over traditional object detectors but also outperforms mainstream query-based detectors when the number of decoder stages is limited to two. As one of the few query-based detectors attempting to achieve decent detection results using fewer stages, we hope our findings encourage further investigations into non-cascade query-based detector design paradigms, ultimately resulting in simpler structures, faster inference speeds, and enhanced performance.

Despite the effectiveness of our approach, we must acknowledge some significant limitations. While our method reduces the need for decoder stages in query-based detectors, the improvement in inference speed is not as substantial as anticipated, primarily due to various query-enhancing operations. We are dedicated to addressing these limitations in our future research.

\bibliography{aaai24}

\end{document}